# One mechanism for many mental spaces: a shared router over a value slot in language models

**Oliver Steele**
steele@osteele.com
Tsinghua University
NYU Shanghai

**Jiangtao Wen**
jw9263@nyu.edu
NYU Shanghai

**Yuxing Han**
yuxinghan@sz.tsinghua.edu.cn
Tsinghua University

2026-07-11

**Abstract**

Language builds discourse contexts other than the actual: a painting, a belief, a memory, a hypothetical. Each is a mental space in which the same entity can take a different value, as when a flower is red in reality but purple in a portrait. Formal semantics keeps these contexts apart because their logics differ (modal, temporal, doxastic, depictive). Fauconnier's mental-space theory, by contrast, treats them as one space-building operation. We ask which of these a transformer language model implements, and find a mechanistic version of Fauconnier's unification. The model uses one router/slot format across the inventory: a reusable *value slot* stores attributed content, and a causally manipulable *router* (the *space index*) selects which space is read. A subspace trained with Distributed Alignment Search to control one space type, counterfactual, belief, fictional, or temporal, also controls the others, well above a random floor, on three model families. Belief, which formal semantics marks as a distinct case, is not specially separated. The router is low-rank, composes additively with entity identity, and acts through a few late-layer heads. Two further results show the mechanism drives inference and composes: a subspace trained on a rule-derived conclusion flips what the model infers while dissociating from what it reports, and composing space-builders mints a fresh router over the shared slot. This paper establishes the cross-type generality. A companion paper develops belief in depth, because of its special status in philosophy, psychology, and linguistics (epistemology, theory of mind, and propositional attitude reports).

## 1 Introduction

A capable language model keeps a flower's real color apart from its color in a portrait, its present apart from its past, and what is true apart from what a character believes. We show it does so with a single, causally manipulable *space index*: a low-rank subspace that tags which discourse space a value belongs to. Formal semantics keeps these contexts apart because their logics differ; Fauconnier's mental-space theory unifies them as one space-building operation.

The model implements a mechanistic version of that unification: a subspace trained on one space type controls the others (transfer index 0.71–0.89 across model families, §4.8), while the semantic distinctions survive as a coarse taxonomy inside the shared subspace (§4.11). This is evidence for a shared representational format in LMs, not an adjudication of the semantics of English.

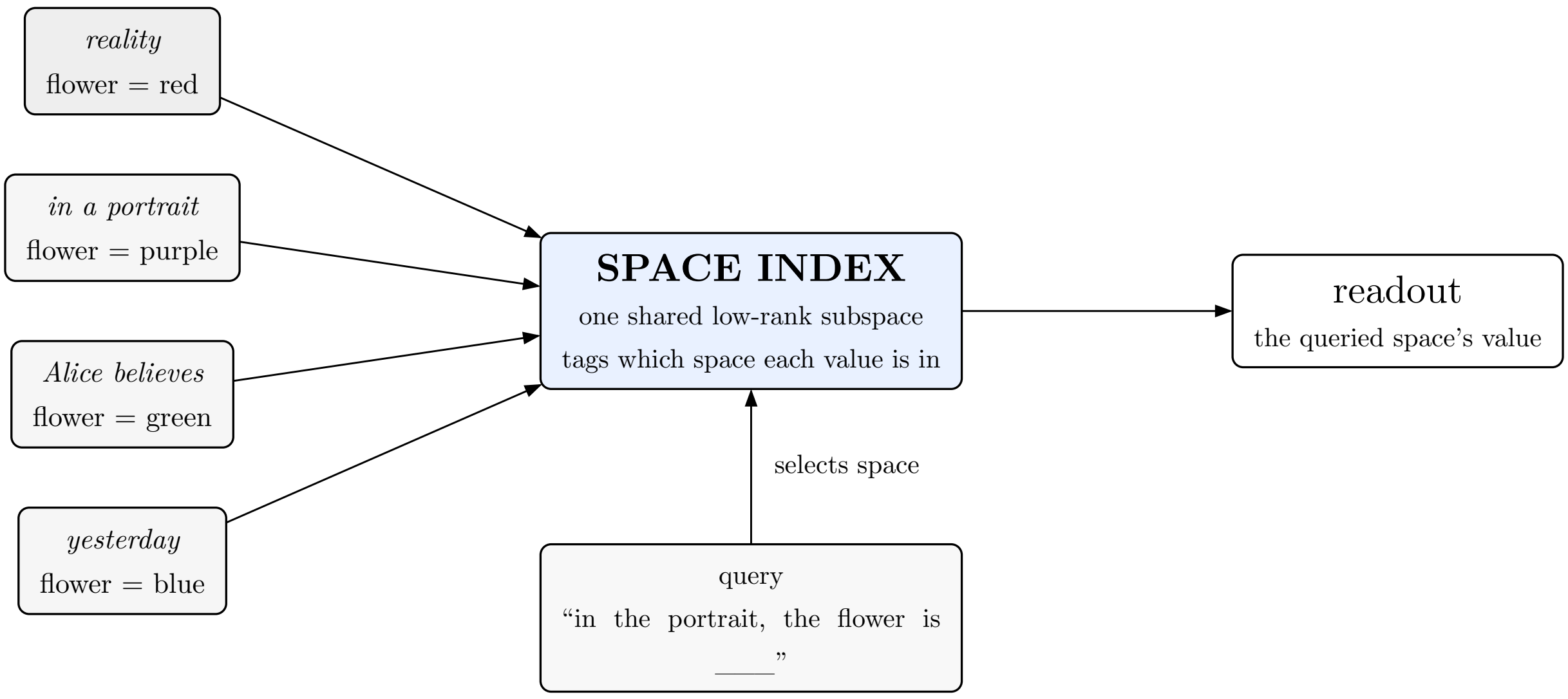

Figure 1: One entity (the flower) takes a different value in each discourse space, and a single shared low-rank subspace, the *space index* (router), tags which space each value belongs to. A query names a space and the router selects its value for the output. We show this one router is causally shared across space types (§4.8), used in reasoning (§4.7), and minted afresh when a sentence builds a new space (§4.10).

Put plainly: whenever a sentence opens a non-actual context, a portrait, a belief, a memory, a hypothetical, the model files each value with a tag for which context it belongs to, and one kind of tag serves them all. The tag is separate from the value it marks, and the same tagging machinery is reused whether the context is a painting, a belief, or yesterday. The rest of the paper localizes that tag, shows it is one shared subspace across context types, and shows the model mints a new one when a sentence builds a new context.

Three questions frame the work. The *computational* question is whether keeping coexisting values apart is merely recovering two color tokens, or whether the model also carries a code for the local context that makes each value valid. Language builds these non-actual contexts with ordinary grammar, *in a portrait*, *Alice believes*, *yesterday*, *if things were different*, and they can coexist around one entity.

The *linguistic* question is how that code is organized. Formal semantics keeps the contexts apart because they license different inferences: counterfactuals and modals quantify over possible worlds, tense shifts the time of evaluation, and attitude reports such as belief are referentially opaque. That opacity, the blocking of substitution between co-referring terms, is what most sharply distinguishes belief from the merely intensional cases. Fauconnier's mental spaces (Fauconnier, 1994) instead treat them as instances of one operation: discourse builds a partial space and binds values there, while preserving links to shared entities. This paper asks which organization a transformer implements.

The *mechanistic* question is whether the local context is itself represented, and if so how, as a dedicated code with its own vehicle, or only implicitly, in the values it governs. Entity-tracking work supplies the vocabulary. Language models attach low-rank binding and ordering IDs to entities, positions, and attributes (Feng and Steinhardt, 2024; Dai et al., 2024); Prakash et al.'s belief-tracking system (Prakash et al., 2026) instead uses ordering IDs and lookbacks to bind character-object-state tuples. The first line keys entities or order slots; the second keys character-object pairs to state values. Neither tests whether the grammar of non-actual reference writes a shared index for the discourse space itself. That is the object localized here.

The paper's terms are fixed in Table 1 and used in those senses throughout. Each names a low-rank, continuous subspace by its functional role, re-read from context when queried rather than stored (§4.2), not a discrete symbol or register; *binding* is used in the mechanistic-interpretability sense, an in-context identifier attached to a value, not the variable-binding sense of formal semantics.

| Term | Object | Locus | Established |
|---|---|---|---|
| **value slot** | stores an attributed value; agnostic to which space reads it | value token | §4.9 companion (Steele et al., 2026) |
| **router** (= **space index**) | selects which space's value a query reads out; the subspace this paper localizes and manipulates | entity/query token | §4.3–4.8 |
| **binding core** | the low-rank component of the slot shared by asserted and derived filling | value token | §4.9 |
| **output writing** | whether a bound value reaches the logits at all, via late-layer heads; distinct from the router's space *selection* | late layers | §4.5 |

Table 1: Terminology. *Space index* and *router* name the same object; *frame* is the companion paper's term for the two perspectives a value can be read under (a character's belief vs reality), with *space* reserved for the discourse constructs studied here.

At the representational level, the answer to the mechanistic question is yes. Both the base and alternative values decode at a later mention. The space index is a low-rank shared subspace (§4.3), composes additively with entity identity (§4.4), and is written to the output by a few late-layer heads (§4.5). A DAS subspace of rank $\leq 4$ fully controls which value the model emits (§4.6). The model therefore carries a compact key for the discourse context, not just two recoverable color tokens.

That key factors into two objects (Table 1). A *value slot* holds the attributed value and is agnostic to which space reads it. A *router*, the space index, selects which space the value is read under; it sits at the entity/query token and is a different object from the value it routes. The router is the subspace this paper manipulates (§4.6–4.8), and the unification claim is that it is *one* router shared across space types, not that the mechanism is unitary.

The main contributions concern what this key does:

- The space index is *causally shared* across the space types formal semantics separates: a DAS subspace trained on one type controls the others (§4.8), its binding core is filled either directly or by derivation across types (§4.9), and the belief builder's tag is not specially separated from the rest of the builder inventory (§4.11).

- A value used for downstream reasoning is *causally separable* from the value the model reports, through distinct subspaces in a single-space rule task (§4.7).
- *Construction is itself a binding operation*: composing space-builders mints a fresh index over the shared value slot (§4.10).

Companion work establishes the slot/router factorization for the belief space: a value-token subspace transfers almost fully across the belief and reality frames, while query-token routers dissociate cleanly (Steele et al., 2026). That paper develops the belief case in depth. This paper is the cross-type generality study: the shared router across builder types (§4.8), the reasoning dissociation (§4.7), and construction (§4.10).

# 2 Related work

**Binding, ordering IDs, and entity tracking.** Feng and Steinhardt (2024) show transformer LMs solve in-context entity-attribute binding by attaching *binding ID* vectors to entities and their attributes; the IDs occupy a continuous low-rank subspace that transfers across tasks. Dai et al. (2024) sharpen this representationally: a PCA-localized low-rank *Ordering-ID* subspace in the entity activations encodes the order of entity and attribute, and patching along it causally rebinds the entity. In a sentence such as "The cup is blue and the bowl is green", the binding ID keeps *cup* paired with *blue* rather than *green*. This binding machinery is the mechanistic core of in-context *entity tracking*: the task (Kim and Schuster, 2023) and its circuit, including how fine-tuning sharpens the same positional mechanism (Prakash et al., 2024). Davies et al. (2023) localize variable-binding circuitry directly with desiderata-based binary masks, and Wu et al. (2025) supply the developmental view: a transformer trained on symbolic programs learns to exploit the residual stream as an addressable memory space, with specialized attention heads routing information across token positions. These results supply tests and vocabulary for this paper, especially low-rank subspace localization, additive factorization, and causal subspace control. The target here is different: the index is not the entity, the entity's order, or the entity's attribute, but the discourse space in which an attribute is valid.

RAVEL (Huang et al., 2024) is the closest methodological benchmark: it asks whether an intervention can separate an entity from its attribute value rather than merely moving an entangled feature bundle. We import that standard in two ways. The additive entity $\times$ space tests ask whether space information is represented beyond entity identity, and the base-vs-image DAS-per-binding test asks for the stronger double-dissociation form. The latter is only partial here, because the explicit-frame readout is retrieval-shaped and leaky; this is why the paper rests the causal claim on learned space control and cross-type transfer rather than claiming a RAVEL-complete disentanglement.

**Belief tracking and lookbacks as a contrast case.** Prakash et al. (2026) analyze belief tracking in Llama-3-70B as a *lookback* mechanism: the model binds character-object-state triples via Ordering IDs in low-rank subspaces, with *state* denoting the object's content or value in their CausalToM task, and, when queried, dereferences a pointer to re-read the bound value. If Bob fills a jar with beer and Carla fills a cup with coffee, the relevant triples bind (Bob, jar, beer) and (Carla, cup, coffee); a query supplies the character and object as a pointer to retrieve the state OI. That system is not a space-index account: the believer is one component of the address, not itself a discourse-space tag, and visibility is handled by a separate lookback whose payload semantics remains only partly pinned down. The belief case is developed in companion work, which shows the lookback and the space index are two stages of one mechanism rather than rival accounts (Steele et al., 2026); here belief enters only as one column of the cross-type transfer (§4.8). Earlier, Zhu et al. (2024) linearly decode and causally manipulate self/other beliefs; Bortoletto et al. (2024) show belief representations are structured but brittle to prompt variation and correctable by activation edits. The padding control here follows their brittleness and steering-fixes-behavior results. The present experiments add a different mechanistic object: a low-rank, output-routed, causally compact space index with additive composition and cross-type routing.

**State tracking, world models, and dynamic semantics.** A parallel line asks whether sequence models carry internal representations of the world state their text describes. Li et al. (2021) find entity and situation representations that evolve through discourse and can be manipulated to affect generation; Li et al. (2023a) and Karvonen (2024) show board-state world models in game-trained sequence models; and Li et al. (2025a) analyze mechanisms for tracking an evolving permutation state. Those results concern a single evolving state. The object here is different: a query-position router that selects among several coexisting discourse spaces, including non-actual spaces, and transfers causally across the grammatical builders that introduce them. The formal-semantics comparison is likewise not unique to mental spaces. Discourse Representation Theory models discourse as an evolving representation with discourse referents and embedded structures (Kamp et al., 2011) we use Fauconnier's framing because the empirical question here is whether counterfactual, temporal, doxastic, and depictive contexts share one space-building mechanism or fragment into separate semantic modules.

**Counterfactual reasoning and its representation.** Li et al. (2023b) find that LMs' ability to override world knowledge under counterfactual conditionals is largely driven by surface lexical cues, leaving open whether a counterfactual world is *represented*. This paper answers the representational question: both the base and alternative values are separably encoded at a later mention (§4.1), and the single-space anchors and copy-from-context controls rule out the lexical/surface reading their critique targets. Concurrently, Miller et al. (2025) show that a transformer on a *synthetic* in-context counterfactual-regression task encodes the latent parameter and locates the capability in the attention sublayers. The present target is the natural-language Fauconnier inventory and the *space-keying index shared across builder types*, rather than a single synthetic latent.

**Causal subspace methods.** Distributed Alignment Search, introduced by Geiger et al. (2024) and scaled by Wu et al. (2023), learns an orthonormal subspace, model frozen, optimized by an interchange objective. Here DAS serves as the on-manifold causal handle when single-site hand-built splices fail (§4.6), and is extended to train *per-readout* subspaces (the report-vs-reasoning double dissociation, §4.7) and to test *cross-type transfer* (§4.8). Sutter et al. (2025) caution that causal abstraction with unrestricted alignment maps is vacuous: any network aligns to any algorithm. A subspace meeting an interchange objective need not be the model's mechanism rather than an expressive fit. Several checks guard against that reading. The alignment map is a *linear* orthonormal subspace, not the arbitrary nonlinear map that drives the vacuity. The leakage baseline and random-subspace control (§4.6) measure what a non-mechanistic splice already achieves, so the learned subspace is reported only as the *margin* above it. The subspace also makes predictions a mere fit would not: a *double* dissociation (§4.7) and cross-type transfer (§4.8), neither expected from an expressively overfit direction.

**Retrieval heads and retrieval-shaped computation.** Wu et al. (2024) identify a sparse set of *retrieval heads* that copy-paste from context and govern long-context factuality; the lookback mechanism (Prakash et al., 2026) resembles a binding-specific instance of the same re-read-from-source motif. This paper does not claim retrieval-shaping de novo: "retrieval-shaped" names where the space index sits on this established spectrum. Decodability is local and distance-sensitive (§4.2); the value is re-read from source rather than carried as a static global tag; and a few late-layer heads write it to the output (§4.5). This is a retrieval-head / lookback profile rather than a persistent slot. The same profile has now been documented for in-context task representations: Li et al. (2025b) find that transferrable task representations become active only at particular tokens, just-in-time, even though task identity is decodable throughout the context, so the re-read-on-demand shape is a recurring pattern, not an idiosyncrasy of discourse spaces.

The cross-type transfer in §4.8 also fits a broader pattern in mechanistic interpretability: model components and algorithms can be reused across tasks that look superficially distinct (Merullo et al., 2024). The claim here is the space-specific instance of that pattern, with reuse measured by causal control of which discourse space is read.

# 3 Methods

The experiments read activations from open-weight models run on controlled natural-language stimuli. The primary model is Qwen2.5-3B (36 layers), with cross-architecture replication on Pythia-2.8b (32 layers) and Falcon3-3B (22 layers), three unrelated families. GPT-2 serves only as an implementation sanity check for the intervention machinery and is not a capable model for the task (it applies the rule consequent at chance), so we report no GPT-2 results.

**Stimuli.** Each item gives one entity (a common noun such as cup, cat, or house) two values of a single attribute, color, with four possible values {blue, green, red, brown} and chance 0.25. One value appears in the base world ("In reality, the {entity} is {c}.") and a different value in an alternative space. The two clauses appear in random order, followed by a neutral probe clause ending on the entity ("... Consider the {entity}"). The alternative space is realised by nine builder types, each a fixed template: painting ("In the painting, the {entity} is {c}."), belief ("{Name} believes the {entity} is {c}."), dream, movie, past, yesterday, hypothetical ("If things were different, the {entity} would be {c}."), modal ("Possibly, ..."), and reported speech. We also include an actor↔ character identity connector (real-life vs in-show). A typical painting item is: "In reality, the cup is blue. In the painting, the cup is green. Consider the cup." $N = 240$ items per condition unless noted; entities, colors, and order are sampled per item. Throughout, we use **image** and **reality** (the painting-anchor's labels, retained in figures and tables) as shorthand for the alternative-space and base-world values.

**Stimulus variants and confound controls.** The color stimuli above are the shared default. Several variants depart from it to test whether the code is a space index rather than a lexical, positional, entity-specific, or color-specific artifact:

| Variant | Stimulus bank | Confound or boundary tested |
|---|---|---|
| **Neutral padding** | Color-free filler between the source clauses and the final entity mention. | Local lexical proximity and recency. |
| **Base-only and shuffled labels** | Reality-only controls and shuffled color labels. | Extraction and chance floor. |
| **Two-entity context-copy** | A target and distractor each receive reality and painting colors, using four distinct colors. | Entity-space binding vs. reading which colors occurred in context. |
| **Cross-entity space index** | Two entities share one reality/painting frame; a space probe trained on one entity is tested on the other. | A shared space tag vs. per-(entity, space) roles. |
| **Space vs order** | Clause order and queried space are crossed: reality-first vs painting-first, reality query vs painting query. | A content-keyed space index vs. a first/second positional code. |
| **Property-general anchors** | No-color reality/movie anchors over hair, wealth, height, mood, material, size, temperature, and cleanliness. | An abstract index vs. a color-output axis. |
| **Entity-type anchors** | Person/object entities crossed with shared properties such as size and cleanliness. | Entity-type generality of the index. |
| **Frame-varied builders** | Four surface frames per builder across depiction, belief, dream, movie, temporal, hypothetical, modal, and reported speech. | A discourse-space tag vs. a fixed builder-template regularity. |
| **Capacity and swap errors** | One entity with $K$ coexisting spaces, distinct values, and one query per space. | Capacity degradation and binding-swap errors. |
| **Rule-consequent stimuli** | A color-to-recipient sorting rule with separate color-report and consequent readouts. | Whether the controlled value is used by downstream reasoning. |
| **Frontier-model QA** | Instruction-following versions with expanded color banks and API logprob readout when available. | Checks behavioral scale without claiming activation-level mechanism. |

Table 2: Stimulus variants used to rule out major confounds.

**Linear probing.** At each layer we fit a standardized multinomial logistic regression on the residual-stream activation at the final (entity) token, with 5-fold stratified cross-validation, and report out-of-fold accuracy with a 95% CI for the base and the alternative color *separately*; "decodable" means the CI lower bound exceeds chance. Controls: a base-only probe (confirms extraction) and a shuffled-label probe (floor at chance).

**Steering.** A diff-of-means direction $v = \text{mean(alt)} - \text{mean(base)}$ at the maintained-representation layer, added to the residual stream (the running activation vector each layer reads from and writes to) with residual-norm-normalized magnitude $\alpha$; we read the color continuation. Steering is evaluated against the *distribution* of 20 matched-norm random directions, each given the same $\alpha$-selection procedure (max over the same $\alpha$ grid) as the steer vector; single random draws proved unreliable for this comparison (§4.6).

**Interchange and DAS.** Activation patching follows causal mediation (Vig et al., 2020), causal tracing (Meng et al., 2022), and activation-patching methodology (Zhang and Nanda, 2024). It replaces the entity-token activation (or its component along a subspace) with a donor item's. DAS learns an orthonormal $R \in \mathbb{R}^{d \times k}$ with the model frozen (only $R$ trained; Adam, lr $5 \times 10^{-3}$, $\sim$150 steps, batch 16, $n \sim$160–240) by an interchange-flip objective: maximize the probability of the donor space's value at the readout when $R$'s component of the target is spliced from the donor. We sweep the rank $k$, and baseline against a rank-1 diff-of-means splice and a random-subspace control. The intervention layer is each model's *maintained-representation* layer: Qwen's empirically located flip-rate peak L34, and Pythia/Falcon3′s depth heuristic ($\sim$0.57 of layers, L19/L13). L34 was chosen on the initial Qwen steering/layer scan, before the DAS rank and transfer tests reported below; it is therefore a fixed analysis locus for the DAS results rather than a held-out per-item selection. This sits at or before each model's output-writing locus (§4.5); on Falcon3 the causal locus (L13) precedes the writing peak (L19–21), consistent with the binding being fixed upstream of where it is read out, though we do not over-read a single model's offset.

For cross-type transfer we report a *transfer index*: the off-diagonal flip rate, normalized between the matched random-subspace floor and the within-type diagonal ceiling for that run. Thus 0 means no better than matched random transfer, and 1 means cross-type transfer reaches the source type's own control. The interpretive bar was fixed in advance: index $\geq 0.7$ counts as causal unification; values below but well above random are reported as partial.

**Output writing.** A logit lens introduced by nostalgebraist (2020), with tuned-lens refinement from Belrose et al. (2023), reads each value's logit at every layer, as elevation over the *unstated controls*: the two palette colors mentioned nowhere in the item. Head-level direct logit attribution (DLA; Wang et al. (2023)) attributes the writing to individual attention heads.

All steering/interchange use single-space anchors where the value is not restated adjacent to the readout, to avoid copy-from-context confounds.

# 4 Results

The results come in three parts. The space index *exists* and is low-rank: both the base and alternative values are decodable at a later mention (§4.1), the code is a low-rank subspace (§4.3) that composes additively with entity identity (§4.4) and is written to the output by a few heads (§4.5), though the binding is retrieval-shaped rather than a static slot (§4.2). The index is causally *used*: a learned subspace controls which value the model emits (§4.6), and a distinct subspace controls the value the model reasons with (§4.7). And it is *one mechanism* across space types: a subspace trained on one builder controls the others (§4.8), the index is filled either directly or by derivation (§4.9), and it recovers a coarse semantic taxonomy without giving the belief-builder tag any special boundary (§4.11).

| Result | Qwen 3B | Pythia-2.8b | Falcon3-3B | Larger / Instruct |
|---|---|---|---|---|
| §4.1 coexisting spaces represented | ✓ | ✓ | – | – |
| §4.2 retrieval-shaped binding | ✓ | ✓ | – | – |
| §4.3 low-rank shared subspace | ✓ | ✓ | ✓ | – |
| §4.4 additive with entity | ✓ | ✓ | ✓ | – |
| §4.5 output routing | ✓ | ✓ | ✓ | – |
| §4.6 DAS causal control | ✓ | ✓ | ✓ | – |
| §4.7 report-vs-reasoning dissoc. | ✓ | – | ✓ | – |
| §4.8 cross-type transfer | ✓ | ✓ | ✓ | – |
| §4.9 filling (asserted/derived) | – | – | – | 14B, Mistral-24B |
| §4.10 construction (router minting) | – | – | – | Qwen, OLMo-2, Mistral |
| §4.11 taxonomy | ✓ | ✓ | ✓ | – |

Table 3: Model coverage by result. The core representational, routing, causal-rank, and taxonomy results (§4.3–4.6, 4.11) hold on three unrelated families (Qwen, Pythia, Falcon3); cross-type transfer (§4.8) on all three (partial on Falcon3); the report-vs-reasoning dissociation (§4.7) on the two families that apply the sorting rule (Pythia cannot); the filling result (§4.9) on Qwen2.5-14B and Mistral-Small-24B. The §4.1–4.2 sub-probes are reported on Qwen and Pythia; the construction result (§4.10) is on instruction-tuned models (Qwen 7B/14B, OLMo-2-7B). Behavioral results replicate at frontier scale (gpt-4o, Qwen2.5-72B, Gemini-2.5). ✓ tested; – not separately reported.

**The space index exists and is low-rank**

## 4.1 Coexisting spaces are separably represented

Take the running example: a cup that is blue in reality and green in a painting, queried at a later mention ("In reality, the cup is blue. In the painting, the cup is green. Consider the cup."). Both values remain decodable at that mention, after the intervening text. On Qwen ($N = 240$ per seed, chance 0.25), the out-of-fold probe accuracy across three seeds is 0.96 $[0.95, 0.97]$ for the base (blue) color and 0.94 $[0.94, 0.95]$ for the alternative (green, painting) color, read at each seed's best layer (the selected layer varies mildly across seeds; the accuracies do not). Across the nine builder types, the alternative-space color decodes at 0.89–0.95 at its best mid/late layer, and stays $\geq +0.15$ over chance even when its clause is *not* the most recent. The effect is therefore not pure recency and survives padding; if anything, it is often anti-recency. The same pattern holds for painting, belief, dream, movie, past, yesterday, hypothetical, modal, reported speech, and an actor↔ character identity connector on Qwen and Pythia.

The code is not a per-(entity, space) role. In a two-entity scene, reality might make the cup blue and the bowl red, while the painting makes the cup green and the bowl brown. The painting-vs-reality direction extracted at the cup token also classifies the bowl token (cross-entity probe transfer 0.88–1.00 ~ within-entity, cross-architecture). The model also holds two entities × two spaces at once (retrieval 0.85 Qwen, 0.45 Pythia). The space index is entity-general.

It is also property-general, and so not a color phenomenon. When the attribute varies across heterogeneous dimensions with no color present (hair, wealth, height, material, size), one "in reality" vs "in the movie" direction still marks the space, transferring across property dimensions at 0.83 (Qwen) and 0.89 (Pythia). A property-general direction cannot be the color-output axis. The tag is entity-type-general as well: holding the property fixed, it transfers between person and object entities at 1.00 / 0.99. The space index is an abstract space tag, invariant to the entity, the property, and the entity type.

The tag also survives surface-frame variation, so it is not a fixed-template regularity. With four surface frames per builder, the represented-space readout remains above chance and above shuffled-label controls for all nine tested builders: painting, belief, dream, movie, past, yesterday, hypothetical, modal, and reported speech. Best-layer alternative-space accuracy ranges from 0.69 to 0.84 across builders, while shuffled-label controls remain at 0.25–0.27 (chance 0.25); the corresponding base-space accuracy ranges from 0.55 to 0.89. The control weakens some readouts relative to fixed frames, but it rules out a single repeated builder template as the source of the decodability result (see Limitations).

## 4.2 The binding is functionally real but retrieval-shaped

The binding is entity-specific, but not stored as a well-isolated global slot. By *retrieval-shaped* we mean it is re-read from its source when queried, a retrieval-head / lookback profile (Wu et al., 2024; Prakash et al., 2026) rather than a value held in a persistent slot. In a two-entity control, a target and a distractor each receive a reality color and a painting color. At the best layer the target's colors decode only marginally above the distractor's (target − distractor +0.07 on Qwen, +0.09 on Pythia); the distractor's colors decode at the target token almost as well (*context bleed*), and the advantage decays to ~ 0 within ~ 16 filler sentences. The binding is re-read from its source, not held in a persistent slot. This decodable-throughout-but-active-locally split matches the just-in-time profile Li et al. (2025b) report for in-context task representations, so the retrieval shape is an instance of a documented pattern rather than an isolated caveat.

This leakage does not mean the binding is behaviorally absent. Frontier models answer entity× space queries under load. A suggestive behavioral probe further indicates that *maintenance* (retrieving an asserted binding) and *inference* (deriving a bound value by rule) come apart across models. The magnitude is not used as evidence here, but the split is not incidental: it is the behavioral face of the report-vs-reasoning dissociation localized in §4.7 and of the asserted-vs-derived filling of §4.9, and the companion paper maps its capability axis for belief (Steele et al., 2026).

## 4.3 The space index is a low-rank shared subspace

Per-type "alt-vs-reality" directions do not collapse to one universal axis. Across all three families, they have effective rank ~ 4–5 of 10: one dominant shared component (~ 40% of variance) plus type-specific structure, with cross-type probe transfer ~ 0.60. The result is a space-index *subspace*, not a single hand-built axis. Falcon3 shows somewhat *more* sharing than Qwen and Pythia (off-diagonal cosine 0.46, cross-transfer 0.76, vs 0.29–0.33 / 0.60–0.61), a quantitative family difference rather than a sign reversal.

## 4.4 It composes additively with entity identity

In an entity × space factorial, "entity E in space S" factorises into entity + space main effects (additive $R^2 \approx 0.92$–$0.95$ across the three families), with the two factors near-orthogonal (principal cosine ~ 0.29–0.34). Entity identity and discourse space compose as separable factors. The geometry is not an artifact of subtracting one fixed reality phrase. Re-running the shared-axis and additivity analyses with fixed, varied, and bare base clauses leaves both summaries intact: shared transfer is 0.626 / 0.606 / 0.721, and additive $R^2$ is 0.948 / 0.949 / 0.947 on Qwen2.5-3B.

## 4.5 Both spaces reach the model's own output, via localized heads

The values are not just decodable from hidden states; they reach the model's own output channel. A logit lens elevates both values far above unstated controls in all three families (Figure 2). Head-level DLA localizes this output writing to a few late-layer attention heads: Qwen L31–33, Pythia L16–20, and Falcon3 L19–21, each matching its own model's logit-lens peak. Image and reality use *distinct* mover heads (mover overlap 0 of 8 in every model), and the alternative-space value is the more actively written one (Figure 3).

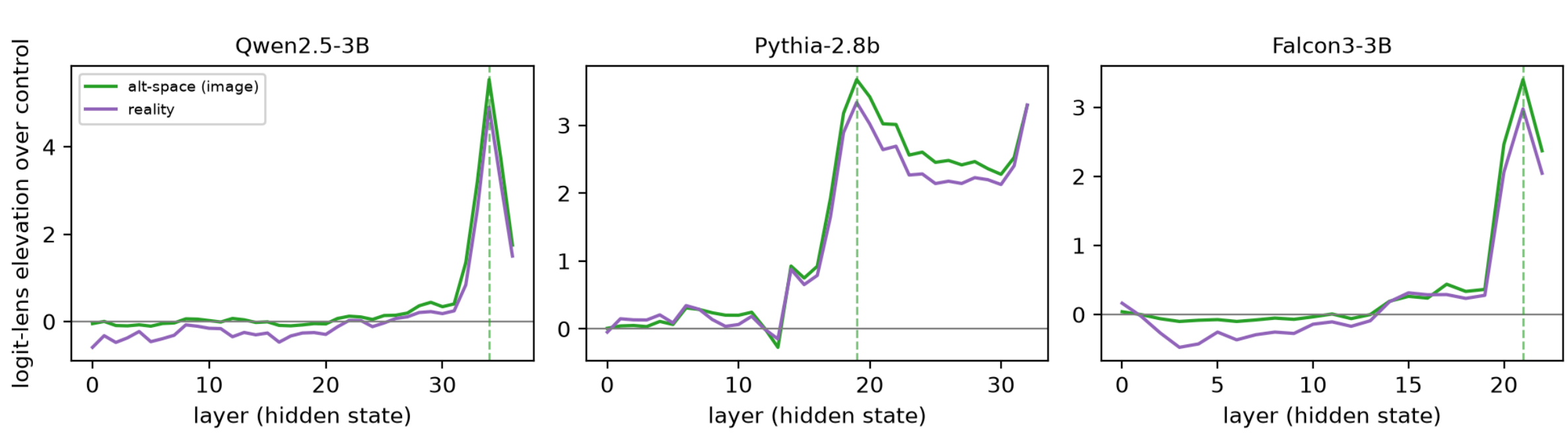

Figure 2: Logit-lens output writing per model: both the alternative-space (*image*) and base-world (*reality*) values rise above an unstated control through the late layers, so both values reach the model's own output channel. The writing locus differs by model; the pattern is shared.

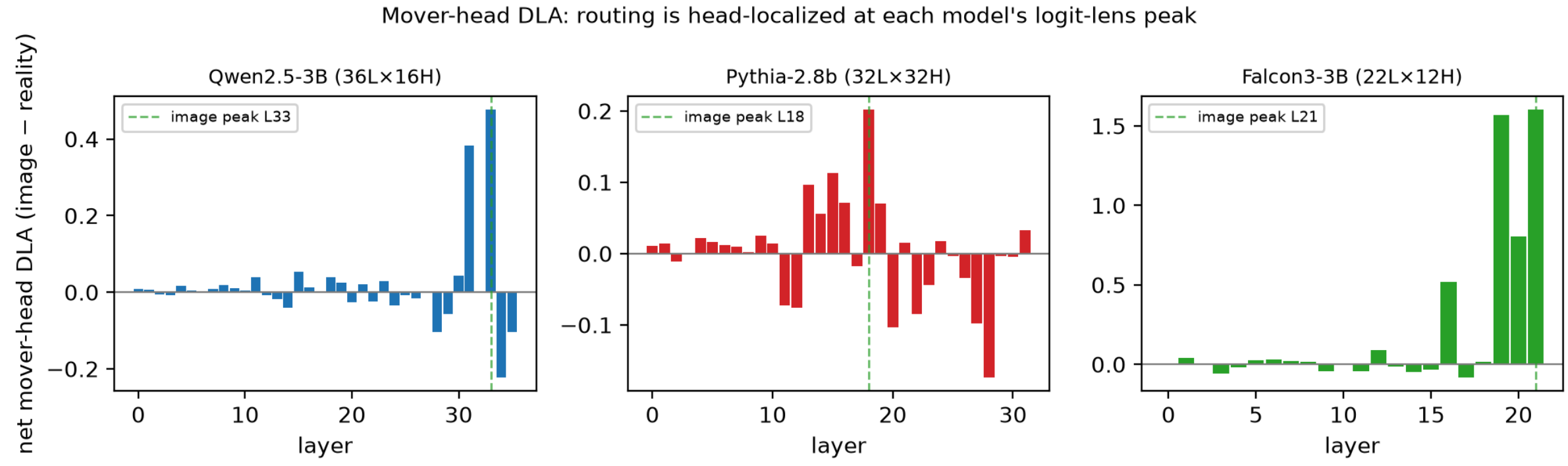

Figure 3: Per-layer net mover-head direct logit attribution (image − reality). Output writing is head-localized at each model's logit-lens peak (dashed line), and the image and reality mover sets are disjoint: the two spaces reach the output through different heads.

**The index is causally used**

## 4.6 Causal control requires a learned subspace

Hand-built steering is indistinguishable from a matched-norm random direction. The steered flip is stable across seeds (0.42–0.50 at the steering layer), but stability is not significance: the 20-draw random-direction band is $[0.16, 0.58]$ (mean 0.40) and contains the steer (0.48). A single draw is an unreliable control here: across three draws at one layer the control flip spans 0.02–0.57 on an identical protocol, a range wide enough to reverse the comparison, so the calibrated band, not any single draw, is the reference. No hand-built direction moves the space→ value choice beyond matched-norm chance; the causal handle appears only when the intervention is learned on manifold. On Qwen, DAS finds a **low-rank learned subspace** that gives full control (rank 1–2 suffices for $\geq 0.9$ on every seed, full control by rank 2–4); across families, the rank is ≤ 4, with rank-1 sufficient on Pythia and Falcon3:

| **Intervention at L34** | **flip → image (held-out)** |
|---|---:|
| random rank-1 (control) | 0.583 |
| diff-of-means rank-1 (hand-built) | 0.583 |
| DAS learned rank-1 | 0.883 |
| DAS learned rank-2 | 0.967 |
| DAS learned rank-4 / 8 / 16 | 1.000 |

Table 4: Causal control of the space→ value choice on Qwen2.5-3B requires a *learned* subspace. Hand-built splices sit at the 0.583 leakage floor; a rank-2 learned subspace nearly suffices and rank 4 gives full control. Every causal number in the paper is the margin over this non-mechanistic floor.

The equal 0.583 values are not a transcription error: in this seed, the rank-1 diff-of-means splice lands exactly at the matched random rank-1 leakage floor. Three seeds give the same reading with small variation: DAS rank-1 mean 0.96 [0.88, 1.00], diff-of-means 0.59 [0.58, 0.62], and random 0.56 [0.53, 0.58] over $n = 200$ held-out items per seed.

Pythia-2.8b (L19) and Falcon3-3B (L13) replicate the result in single-seed runs: DAS reaches flip 1.00 at rank-1 (vs 0.58 and 0.70 rank-1 baselines). The space index is therefore causally as well as geometrically low-rank (causal rank ~ 1–4, matching the geometric rank ~ 4–5) across three architectures (Figure 4, left). Hand-built interventions fail three different ways: additive steering sits inside the random band, the diff-of-means splice sits at the leakage floor, and single-site hand splices go off-manifold or copy from context, while the optimized on-manifold DAS subspace is the working causal handle. The baseline is not zero (0.55–0.58 leakage: a full-magnitude rank-1 splice of the painting-anchor leaks the image color through *any* direction), so the result is the *margin* a low-rank *learned* subspace adds, plus the demonstration that low rank suffices for full control.

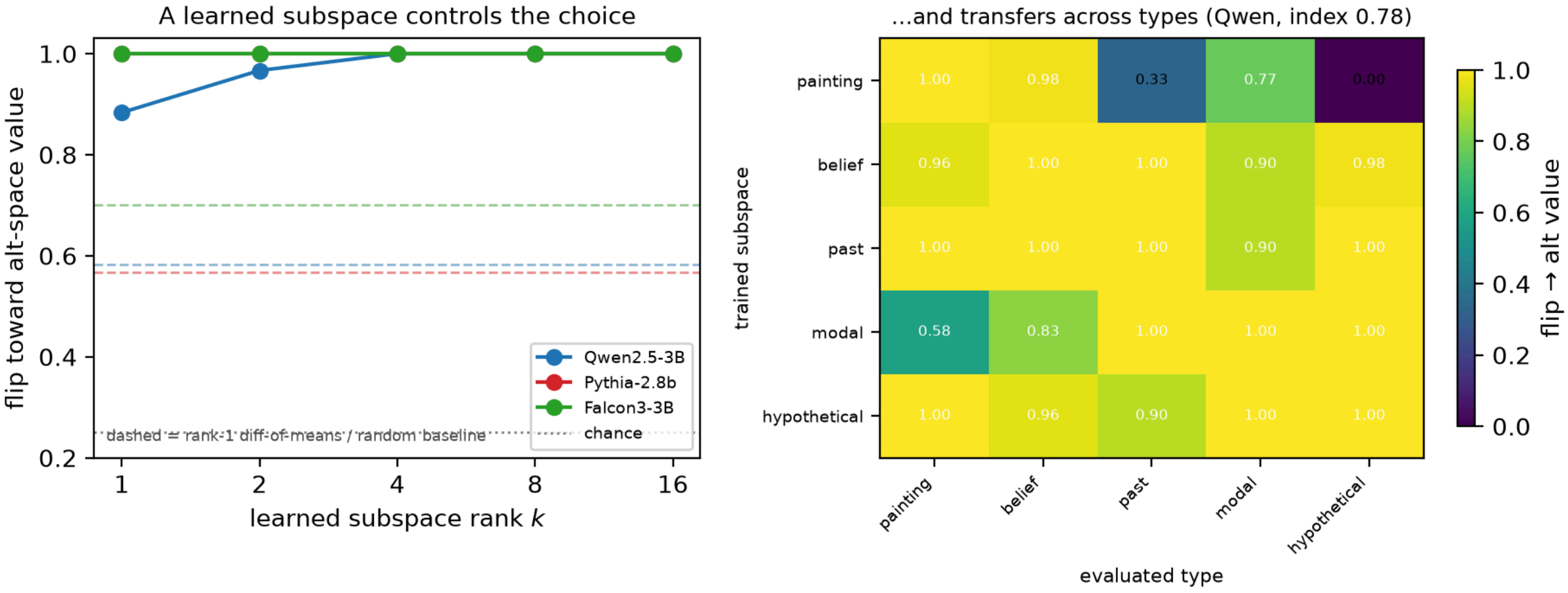

Figure 4: The two central causal results together. *Left*: DAS flip toward the alternative-space value vs learned-subspace rank $k$, per model — a rank ≤ 4 learned subspace gives full causal control, while the rank-1 diff-of-means / random splices (dashed) sit at the ~ 0.58 leakage baseline. *Right*: the same subspace transfers across builder types (Qwen) — each cell is the flip-toward-alt rate when a subspace trained on the *row* type is applied to the *column* type; off-diagonal rates are near the diagonal and above the random floor, with painting the partial outlier as a source.

## 4.7 Report and rule-use values are causally separable

The causal question is whether the value the model *emits* is the same value it uses for downstream *reasoning*. In a single-space stimulus with an explicit sorting rule (blue/green → Anna, red/brown → Ben), Qwen applies the rule perfectly (baseline consequent 1.00). A DAS subspace trained on the *color readout* fully flips the reported color (1.00) yet leaves the rule-derived consequent unmoved (0.00). A second DAS subspace trained on the *consequent readout* fully flips the consequent (1.00) and only weakly perturbs the reported color (0.32):

| DAS subspace | color report | rule consequent |
|---|---|---|
| color-trained | **1.00** | 0.00 |
| consequent-trained | 0.32 | **1.00** |

Table 5: The report-vs-reasoning double dissociation on Qwen2.5-3B: a subspace trained on the color readout fully flips the reported color but leaves the rule-derived consequent unmoved, and the consequent-trained subspace shows the reverse pattern. The value the model emits and the value it reasons with are controlled by distinct subspaces. Each flip rate is a single-seed DAS run ($n = 240$ held-out items); the dissociation's robustness rests on cross-architecture replication (Falcon3, this section) and the measured inter-subspace angle rather than a per-seed CI.

This single-space result separates report from reasoning. The value the model emits and the value it applies a rule to are controlled by distinct subspaces, even though both depend on the same bound color. The consequent *is* reachable from the color token (ruling out "the patch cannot reach reasoning at all"), but only through a subspace distinct from the report one. This is the causal mechanism behind the behavioral maintenance-vs-inference split (§4.2): controlling the stored / reported value is not the same as controlling the inference computed from it, because they occupy different directions.

The dissociation replicates on Falcon3 (report subspace flips the color 1.00 and the consequent 0.49; reasoning subspace flips the consequent 1.00 and the color 0.29), and the two subspaces are substantially non-aligned by principal angle (mean cosine 0.11 on Qwen, 0.17 on Falcon3), so the dissociation rests on a measured separation of subspaces, not the cross-effect asymmetry alone. Falcon3 is therefore a geometric dissociation with a sizable report-to-consequent cross-effect, not the clean double dissociation seen on Qwen. On Pythia-2.8b the test is uninterpretable: the model does not apply the rule (baseline consequent near chance), so it neither confirms nor refutes.

This dissociation is the representational counterpart of the chain-of-thought faithfulness literature, which shows behaviorally that what a model says can fail to reflect the computation that produced its answer (Turpin et al., 2023; Lanham et al., 2023). The present result mechanizes that gap, and sharpens it: the reported value and the rule-applied value are separated causally, by measured distinct subspaces over the same bound token, rather than inferred from output inconsistencies under prompt perturbation. A faithfulness failure is, on this account, an intervention on one subspace that leaves the other untouched.

**One mechanism across space types**

## 4.8 The shared mechanism is causally general; only the rank-1 hand-built axis is type-specific

Concretely, one cross-type transfer proceeds as follows. We learn a low-rank subspace that controls the value in a *hypothetical* space, the one that swaps the color in “if things were different, the flower would be white.” We then apply that same subspace, unchanged, at the query token of a *belief* item (“Alice believes the flower is green”) and a *temporal* one (“yesterday the flower was blue”), and it moves their values too. The machinery learned on one space type keys the others: that single transfer is the generality claim in miniature, and the rest of this section makes it quantitative and shows that a random subspace of the same size cannot do it.

A hand-built rank-1 direction makes the code look more type-specific than it is. The dominant shared axis, or another type’s diff-of-means, steers a target space markedly worse than its own type-specific direction (+0.15 / +0.07 vs +0.39 over single-draw random controls; margins of this kind are draw-sensitive, §4.6, which is itself a reason not to rest causal claims on hand-built steering). A *learned, on-manifold* DAS subspace tells the causal story: a subspace trained to control one builder type, here the *hypothetical* builder, controls the *others*. The transfer uses a rank-8 subspace, and its capacity is controlled by the matched random baseline: a *random* rank-8 subspace flips only 0.28–0.38, far below the trained cross-type transfer, so the effect is the trained alignment and not the dimensionality of the subspace. The mechanism is also rank-4 compact: a rank-4 learned subspace matches the rank-8 transfer within 0.06 on every family (rank-4 index 0.79 / 0.82 / 0.71 on Qwen / Pythia / Falcon3), so a rank-4 subspace already controls all builder types. The minimal rank is family-specific: on Qwen and Pythia the transfer degrades at rank 2 (− 0.14 / − 0.22), placing the causal rank near 4 and matching the geometric rank of ~ 4–5 (§4.3); Falcon3′s mechanism compresses further still (rank 2 as good or better, 0.75), its causal rank falling below its geometry. The rank-8 cross-type flip matrix reaches the within-type ceiling on Qwen and Pythia, and rises more partially on Falcon3. Three seeds give stable transfer on all three families (Table 6). Falcon3 is lower than the other two, but its off-diagonal transfer stays well above the matched random floor.

| model | within-type | cross-type | transfer index |
|---|---|---|---|
| Qwen2.5-3B | 1.00 | 0.81–0.85 | 0.73 [0.67, 0.78] |
| Pythia-2.8b | 1.00 | 0.915–0.916 | 0.885 [0.882, 0.889] |
| Falcon3-3B | 1.00 | 0.792–0.843 | 0.711 [0.662, 0.741] |

Table 6: Cross-type transfer of the space router: a rank-8 DAS subspace trained on one builder type, applied unchanged to the others. The transfer index normalizes between the matched random floor (0) and the within-type ceiling (1); the interpretive bar of 0.7 was fixed in advance. Each cross-type cell uses $n = 160$ held-out items; bracketed ranges are over three seeds per family. These are the transfer-index values cited throughout the paper.

The space-indexing *mechanism* is causally shared across modal, temporal, doxastic, fictional, and counterfactual builders (Figure 4, right). What is type-specific is the *rank-1 hand-built axis*, not the mechanism, the same on-manifold-vs-hand-built lesson as DAS (§4.6) versus the failed single-site splices (§4.2). The one consistent residual is the fictional builder (painting), a partial outlier as a *source* across the tested architectures.

The generality extends beyond the trained builders. A subspace trained on a reference type also controls builders introduced only afterward, including prospective tense: a past-trained subspace flips the *future* space at 0.81, 0.90, and 0.83 across the three families (single seed per family), well above its 0.27–0.33 random floor. The one mechanism covers a broad inventory, modal through temporal.

## 4.9 The binding core is filled directly or by derivation, across space types

The transfer results show that one router tags which space a value belongs to. A separate question is how a value gets into a space. A value can be given directly, an *asserted* value, or withheld and inferred from other information, a *derived* value. This section accordingly works at the answer-relevant *value* token, the slot locus, rather than the query-token router manipulated so far. The two loci come from the companion factorization: at the value token, a subspace trained on the belief readout moves the reality readout at 89–95% of its own range (three architectures, five layers, demonstrations decorrelated from the test templates), while the query-token routers dissociate (cross-leg transfer below a 0.30 analysis bar, top principal cosine 0.19–0.28) (Steele et al., 2026). Companion work also establishes both filling cases for belief: an asserted belief binds directly, while a belief that must be inferred from what a character could see is computed by a visibility-gated lookback and written into the same binding subspace, so asserted and derived belief share a low-rank core (Steele et al., 2026). The same filling structure holds in two further space types.

For the counterfactual and depictive spaces we separate an asserted value, given directly, from a derived value, selected by which antecedent branch or which depicted source is realized. We train a DAS subspace to control each readout on Qwen2.5-14B and measure cross-transfer with the same interchange protocol, here at the answer-relevant *value* token, the shared-slot locus, not §4.6's query-token site (see Discussion for the two loci). In both spaces the two subspaces share a low-dimensional core, three aligned directions, and transfer across it well above the zero baseline. With the belief case the pattern holds across three space types:

| space | asserted → derived | derived → asserted |
|---|---|---|
| belief | 0.74 | 0.79 |
| counterfactual | 0.33 | 0.86 |
| depictive (painting) | 0.64 | 0.51 |

Table 7: Cross-transfer between the asserted-value and derived-value subspaces within each space type, on Qwen2.5-14B: the fraction of a subspace's own controllable range recovered on the other readout (0 = random-subspace baseline, 1 = full control of its own readout). All three share a low-rank binding core, which for the counterfactual and depictive spaces also replicates on Mistral-Small-24B. The asserted/derived asymmetry is not architecture-stable (see text) and is not read as a property of a space.

In every space a directly-given and a derived value draw on one binding core, so the core is filled either way rather than by a separate device per source. The shared core is the robust, replicated finding: it holds across all three space types on Qwen2.5-14B and replicates on a second architecture for each, belief on Mistral-7B in companion work and the counterfactual and depictive spaces on Mistral-Small-24B. The asserted/derived asymmetry's size and direction vary by architecture (the counterfactual's strong Qwen2.5-14B asymmetry turns symmetric on Mistral-Small-24B), so we read only the shared binding core, not the asymmetry, as a property of a space.

The filling route itself carries a cross-space code. Training the route subspaces in one space and applying them in another, a subspace transfers to the *same* route more strongly than to the other (route-matched 0.88 vs route-mismatched 0.72 on Qwen2.5-14B, over a near-zero random floor). The route-specificity is carried almost entirely by the derived side: the belief-derived subspace controls the depictive-derived readout near-totally (0.996), while asserted subspaces transfer near-equally to both routes (0.77/0.75). This matches the subspace geometry, in which asserted and derived subspaces share only the generic binding core while derived subspaces show additional alignment *across* space types: derivation, not assertion, leaves a space-general signature in the binding. The geometry holds at 7B and 14B (at 7B over the six saved subspaces); the causal route-transfer result is single-model (Qwen2.5-14B).

## 4.10 Building a space mints a distinct router over the shared value slot

Space-builders do more than select among existing spaces: they *build* them. The basic operation is *composition*: nesting one builder in another, as "Alice used to believe the cup was blue" builds a past × belief space over its base, the present belief. A space, on this paper's account, is a binding index (§4.3–4.8), and that index factors into a value slot and a router at distinct tokens (see Discussion; (Steele et al., 2026)). Composition then makes a sharp prediction: a built space should *reuse* the base's value slot while *minting* its own router, rather than collapsing onto its base or carrying a wholly separate representation.

**Language models track a broad range of built spaces, with one revealing exception.** Before asking how a built space is indexed, we check that the model represents it at all. Across a battery of construction types held over a shared object against its base, temporal ("used to believe"), reported speech ("once said"), depiction tense ("the painting used to show"), negation ("no longer believes"), belief nesting ("Y thinks X believes"), and counterfactual conditionals ("would believe ... if the lights were on"), a capable model reads the built space's value at ceiling (compound and base both ≈ 1.00) with no cross-frame bleed (≈ 0), holding each built space distinct from its base. The lone exception is *modal possibility*: "$X$ might believe the cup is blue" is not separated from "$X$ believes it is red" (modal compound 0.04–0.62, with 0.4–0.96 of the mass on the *actual* color), the model collapsing the possible onto the actual rather than opening a distinct modal space. Modal aside, the construction battery is behaviorally interpretable, which licenses the mechanistic question (is a built space a fresh index over shared machinery?) for the types the model represents. On smaller open models in bare completion, some built-space readouts, notably past-belief, are weaker than at frontier scale. The mechanistic probes below therefore gate on a clean readout per model, and use the chat/split framing where the completion readout does not clear it, as the companion paper does for second-order belief (Steele et al., 2026).

**A composite space reuses the base's value slot but mints its own router.** We test the past $\times$ belief compound at the two loci the belief mechanism factors into: the *value token*, where the bound value lives, and the *query token*, where the space is selected. At the value token, a DAS subspace trained on the compound's (past-belief) value transfers to the base's (present-belief) value at 0.90 (present$\rightarrow$ past 1.00, past$\rightarrow$ present 0.91) over a random floor of 0.03, sharing a three-dimensional core: the compound *reuses* the base's value slot, at the magnitude of the belief$\leftrightarrow$ reality slot-sharing in the companion paper. At the query token the compound's router is *distinct*: a past-belief router and a present-belief router each flip only their own space's value (diagonals 1.00), transfer nothing to each other (cross $\frac{0.00}{0.00}$), and are near-orthogonal (top principal cosines $\leq 0.26$), reproduced across two architectures (Qwen2.5-7B/14B and OLMo-2-7B-Instruct) and seed-stable.

The same two-part signature, distinct routers over a shared slot, replicates in three further compounds: a nested depictive frame in both orders (a painting of a photograph, and a photograph of a painting) and a mixed belief-depictive compound (a painting inside a belief). Table 8 collects the four cases with their readout gates.[1] Every confirmed cell clears its gates and shows zero anchor-color injection; the remaining cells are readout-gated gaps, not negative mechanism results. Painting-in-belief replicates on three model families and each mirror depictive order on two. Composition therefore reuses the slot and mints the router: a built space is a fresh router over shared value machinery.

[1]The painting-of-photograph readout required prompt repair before its mechanism runs; the repair criterion, fixed in advance, was that all three explicit frame readouts clear 0.80 clean accuracy. The pre-repair nested readout failed that gate.

| Compound | Families | Router mint (cross ratios) | Slot reuse (transfer / floor) | Readout gates |
|---|---|---|---|---|
| past × belief | Qwen2.5-7B/14B, OLMo-2-7B-Instruct | 0.00 / 0.00; cos ≤ 0.26 | 0.90 min / 0.03 | present 0.98 / past 0.83 |
| painting of photograph | Qwen2.5-14B-Instruct + 1 family | 0.00 / 0.00 | 0.93 / 0.02 | 0.844 / 0.941 / 0.816 |
| photograph of painting | two families | 0.00 / 0.02 | 0.885 / 0.000 | 0.991 / 0.994 / 1.000 |
| painting in belief | Qwen2.5-14B-Instruct, OLMo-2, Mistral-Small | 0.00 / 0.019 | 0.968 / 0.010 | gate-cleared |

Table 8: Composition reuses the slot and mints the router, across four compound types. *Router mint*: cross-transfer ratios between the compound's router and its base's (own-space diagonals 1.00, zero anchor-color injection everywhere). *Slot reuse*: value-token cross-space transfer over its matched random floor; the past × belief slot-reuse magnitude is on Qwen2.5-14B-Instruct. *Readout gates*: clean-accuracy checks cleared before each mechanism run (bare / inner / nested where applicable). The past × belief router runs use split evaluation with $n_{\text{test}} = 60$ per seed; the nested depictive and mixed runs use $n = 320$, $k = 8$, confirmed as a cross-architecture mechanism only after the readout gates clear.

**Construction mints a router.** Building a composite space is, mechanistically, minting a new router over the model's shared value slot: the constructive form of the binding-index analogy that §4.3–4.8 established for primitive spaces. Space-building, Fauconnier's defining operation, is thus itself an operation on the binding machinery, not a separate faculty. The same signature in nested depiction and in the mixed painting-in-belief compound argues against a belief-only account of composition, and the confirmed cells replicate across model families.

Three limits scope the claim. First, the router-mint leg reproduces across two architectures, cross-size and seed-stable, but the slot-reuse magnitude is on Qwen2.5-14B alone, though the frame-agnostic value slot it draws on is itself cross-architecture (companion work, incl. OLMo-2). Second, these methods fix the *existence* of a distinct router over a shared slot, not the *compositional structure* of the built router: a discriminatively-trained router isolates the modifier axis by construction, so whether the built router has an independently recoverable constituent structure (belief + tense) is left open. Third, the evidence spans different loci and model regimes: primitive-space routing is tested at the query token on 3B base models, value-slot reuse at the value token on 14B/Mistral-24B substrates, and construction largely on instruction-tuned models (with a Mistral-Small-24B base replication) where the compound readouts clear behavioral gates. The construction result should therefore be read as a gated mechanistic signature, not yet a base-model comparison.

The causal router is also not tied to one painting sentence template. A painting router trained on the canonical frame transfers to "shows", "depicts", and "according to" paraphrases on Qwen2.5-3B, with patched image readout 0.982–1.000 and zero anchor-color injection. This control addresses the narrow fixed-template concern for the painting causal leg; fully naturalistic text remains open.

## Behavioral calibration: capacity and inference under load

The mechanistic tests above are on open-weight models, but the behavior they explain is not a small-model artifact. Three behavioral batteries calibrate the scope. First, in a stated-binding task, served frontier models maintain all entity × space bindings through $K = 8$ objects (16 queried cells) with no swap errors. Second, when the painting values must be derived from a rule rather than stated, maintenance and inference separate: the strongest models still apply the rule under load, while smaller models can retain stated bindings yet fail the derived values. Third, a capacity-isomorphism test on the open 3B families does not show the entity-binding-style degradation curve as more coexisting spaces are added.

| Battery | models | result | interpretation |
|---|---|---|---|
| Frontier frame/rule/capacity | gpt-4o, Qwen-72B, Gemini-2.5 | all 1.00; K1–K8 1.00; swap 0.00 | behavior is clean at scale; API access does not expose activations |
| Stated binding under load | GPT-4.1, Claude, Gemini, Llama-70B | K8 acc 1.00; full-bind 1.00; swap 0.00 | stated entity × space maintenance is intact |
| Derived rule under load | same served set | K8 derived 1.00 to 0.90; Llama-8B 0.31 | inference, not maintenance, is the capability axis |
| Open-model capacity | Qwen / Pythia / Falcon3 3B | Qwen 0.95/0.93/0.85/0.91 for K1–K4 | no monotone entity-binding capacity curve |

Table 9: Behavioral calibration from EXP-022, EXP-001e, and EXP-014. Frontier models are at ceiling on binding, rule application, and K≤8 capacity; over K≤4 the open 3B models show no entity-binding-style capacity curve (Qwen flat-high, Falcon3 flat-mid, both unsaturated; Pythia sits at its 4-way chance floor, so it cannot test degradation). The hard behavioral axis is deriving values under load, not retrieving stated values.

### 4.11 The shared subspace recovers a coarse taxonomy without a special belief boundary

Within the shared space-index subspace, the builder types cluster by semantic class: grouping the nine non-identity types into attitude-report builders (belief, reported), intensional/modal (hypothetical, modal), temporal (past, yesterday), and fictional (painting, movie, dream), the within-class probe transfer (0.89–0.92) far exceeds between-class (0.53–0.76) on all three architectures. This is the geometric form of the residual type-specific structure. But the boundary formal semantics treats as deepest is not specially marked by this builder-tag probe. Belief reports are hyperintensional, referentially opaque contexts, where substitution of co-extensional terms can fail (Cresswell, 1975; Quine, 1956). In the model's subspace, belief is no more separated from the counterfactual (hypothetical) space than from any other class: the belief-counterfactual transfer sits *at or above* the between-class baseline on every model, never below it. We read this as evidence, not proof, of "no privileged status," but the null is informative: the method does resolve class structure of the relevant magnitude. The within-class vs between-class separation it detects is +0.16 to +0.37, so a hyperintensional boundary of comparable size would be visible, and is not. The null is, moreover, what the theory under test predicts rather than merely tolerates: in Fauconnier's own treatment, referential opacity and de re/de dicto ambiguities are not properties peculiar to attitude reports but consequences of the general Access Principle governing cross-space reference (Fauconnier, 1994). A belief tag no more separated than any other space tag is the expected signature of that account, so the missing boundary is positive confirmation of the specific theory, not only a failure to find the formal-semantic one. The causal cross-type matrix gives the same answer on the smaller five-builder set: belief↔ modal/hypothetical transfer is above the ordinary between-class baseline on Qwen, Pythia, and Falcon3 (+0.08, +0.03, +0.09). The causal residual is instead the depictive source: painting-trained subspaces transfer less well than non-fictional sources on all three models. Two conditions per class bound the precision.

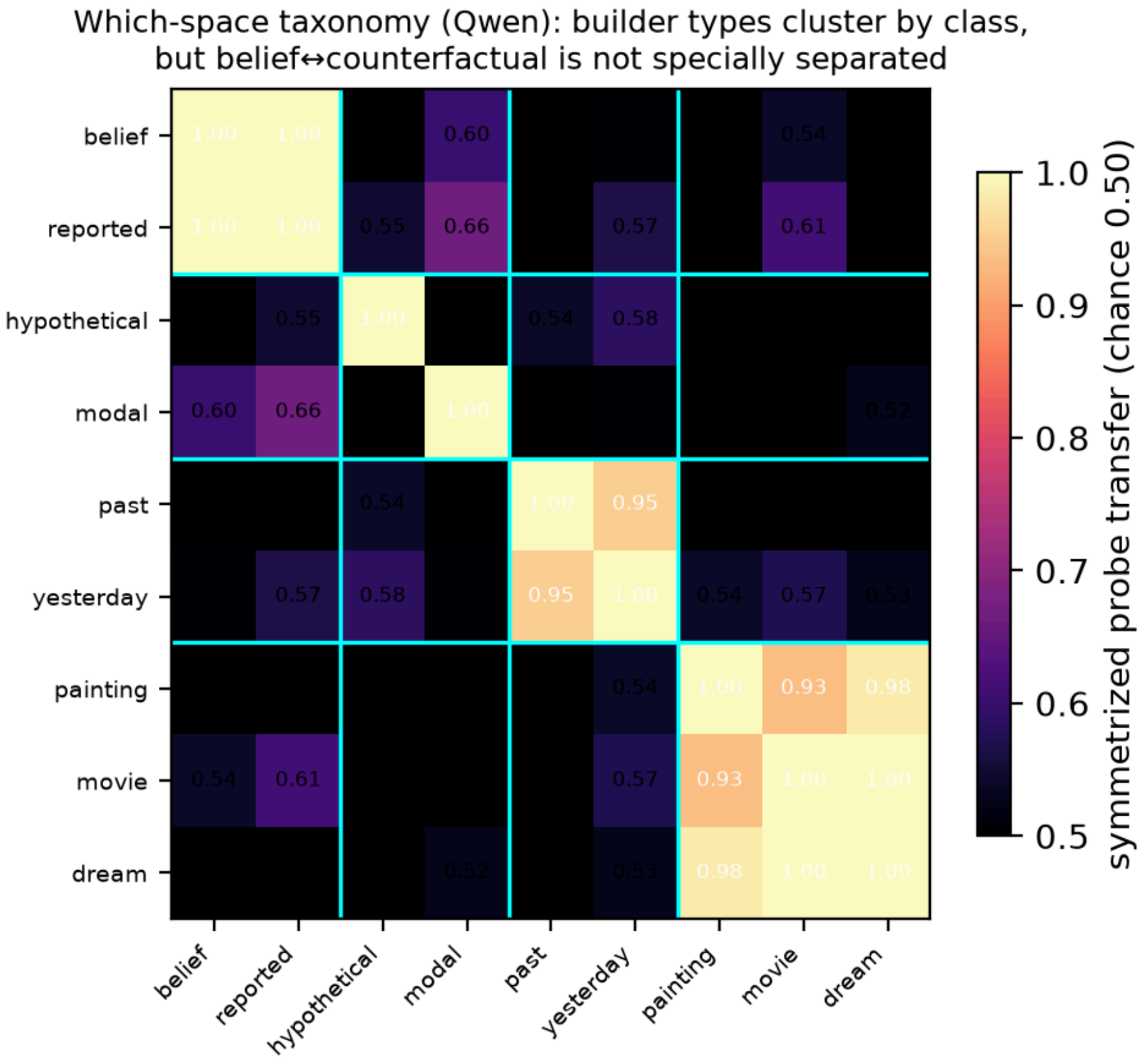

Figure 5: Which-space probe-transfer between builder types (Qwen), ordered by semantic class (cyan lines mark class blocks). On-diagonal blocks are brighter (within-class clustering); but the attitude-report block (belief, reported) is no brighter against the intensional block (hypothetical, modal) than against any other: the belief-builder tag is not specially separated from counterfactual.

# 5 Discussion

The space index is a discourse-space variable. It is neither entity binding nor Prakash-style character-object-state belief tracking. Its low rank, additivity with entity identity, output routing, and causal compactness make contact with binding-ID work (Feng and Steinhardt, 2024; Dai et al., 2024), but the localized object is different. The contribution is the space key itself and its behavior across the semantic inventory.

**A format shared across semantically distinct phenomena.** The builder types we probe are, in formal semantics, *different objects*: counterfactual/modal spaces are possible-worlds modality (Stalnaker, 1968; Lewis, 1973); belief is a hyperintensional, referentially opaque doxastic context (Quine, 1956; Cresswell, 1975); "in the painting / the movie" is fictional-depictive; past/yesterday is tense, not modality at all; and the actor↔ character connector is a pragmatic-function mapping. These are intensional-semantic distinctions (Montague, 1973), and standard accounts keep them apart because their logics differ. Fauconnier's mental-space theory unifies them as one space-building operation, a cognitive-linguistic claim the formal tradition has largely declined to adopt. The model implements this unification causally: across modal, temporal, doxastic, and fictional builders, the space index appears as one shared low-rank subspace (§4.3), additive composition with entity identity (§4.4), and the same output-writing profile (§4.5). A DAS subspace *trained on one builder type controls the others* (Table 6). This is one shared causal mechanism, beyond shared geometry. Prior entity-binding and belief-tracking work do not test this: one stays within entity tracking, the other within a character-object-state belief-tracking task.

The unification is not flat. A coarse semantic taxonomy survives inside the shared subspace: builder types cluster by class, and the fictional builder is a partial causal outlier. But the taxonomy gives the belief-builder tag no privileged status (§4.11). Belief is no more separated from counterfactual than from any other space class, and the null survives a sensitivity check; it is also the pattern Fauconnier's Access Principle predicts, opacity being on his account a general fact about cross-space reference rather than a special property of attitudes (§4.11). This does not test referential opacity itself: there is no co-referential substitution or de-re/de-dicto contrast in the stimuli. The result is instead a **causally shared space-index mechanism with a coarse semantic overlay**: a mechanistic version of Fauconnier's "one operation," with the semantic distinctions surviving inside the shared format rather than requiring separate representational families.

**A distinct space index, not a relabeling of entity binding or belief tracking.** Three checks rule out stronger identifications with familiar binding mechanisms. First, the space-index subspace is not the color-output subspace: the space direction is property-general, marking the space across non-color attributes (§4.1). Second, the space types do not share their output-writing *heads*. The per-layer writing profile is common (§4.5), but the heads that write each space's value overlap only partially and inconsistently across architectures, so the "same output-writing profile" above means the same layer band, not the same heads. Third, space binding does not degrade with the number of coexisting spaces the way entity binding degrades with entities: accuracy is flat from one to four coexisting spaces, with no capacity curve. Entity binding is therefore a comparison class for the controls, not an identity claim. A further comparison is the linear *truth direction* recovered by unsupervised probing (Burns et al., 2023) and mass-mean intervention (Marks and Tegmark, 2024): the base (reality) frame is adjacent to it, but the space index is the richer object: it selects among many non-actual frames, not a binary true/false, and is property-general (§4.1) rather than a factuality axis.

**A value slot and a router, not one "index".** The causal handle localized here sits at the entity/query token, the position that selects *which space's* value a readout retrieves. Companion work on the belief instance separates this from the value token itself (Steele et al., 2026). At the answer-relevant value token, a subspace transfers almost fully *across* frames (belief ↔ reality, 89–95% of its own-frame effect; three architectures, five layers, two demonstration conditions): a generic value-binding slot, agnostic to frame, space type, and filling route. At the query position, low-rank subspaces select *between* frames and dissociate cleanly (cross-frame transfer ≤ 0.3 of the own-frame effect, near-orthogonal routers, zero value-injection under a color-decorrelated control; Qwen, Mistral, and OLMo-2). The space-indexing machinery, in other words, factors into the two objects of Table 1: a shared slot that carries the bound value, and a router that selects which space's value a query reads. The subspace this paper manipulates lives at the router locus, which is why it controls the space→ value *choice* rather than the emitted color as such. A color-decorrelated control confirms this at the space-index instance too: with a painting-frame anchor whose color differs from the probe's, the output flips to the probe's *own* value with injection ≤ 0.13 and cross-frame ratios ≤ 0.08, on Qwen2.5-3B (the substrate of this paper's DAS results), 7B, and 14B. The dissociation and the cross-type sharing of §4.8 are compatible rather than in tension: the sharing is among *marked-space* selections, while the dissociation is between a marked space and the default frame; whether two marked spaces' routers dissociate pairwise is open.

**Construction is a binding operation.** Building a composite space is itself an operation on the binding machinery (§4.10): the model *reuses* the base's value slot while *minting* its own router, so a built space is a fresh router over shared value machinery: the constructive form of the binding-index analogy, just as an entity binding-ID is minted for a new entity without carrying that entity's content. Whether the built router has an independently recoverable constituent structure is left open there.

**Report and reasoning are causally separable.** The report-vs-reasoning double dissociation (§4.7) is the novel *mechanism*: a subspace controlling the value the model emits is distinct from the one controlling the value it reasons with, giving the behavioral faithfulness gap (Turpin et al., 2023) a causal substrate (§4.7). The persistent qualifier is that the binding is **retrieval-shaped**. Decodability is local (§4.2), single-site hand interventions fail, and the value is re-read from source rather than carried as a static global tag, resembling the lookback profile of Prakash et al. (2026). That retrieval shape makes the dissociation possible. The report and rule-application paths re-read the bound value through different features, so an intervention optimized for one readout need not move the other. It is also why the causal handle had to be an *optimized on-manifold subspace* (DAS) rather than a raw splice. Earlier hand-built consequent-flip designs failed under adjacent color restatement, which allowed the model to copy from context instead of using the intervened value.

# 6 Conclusion

A capable language model represents a discourse space with a shared low-rank router over a value slot: the router selects which space a value is read under, composes additively with entity identity, and is causally controlled at rank ≤ 4. That router is one mechanism across the space types formal semantics keeps apart, a subspace trained on one controls the others; it is used by downstream reasoning through a subspace distinct from the one that reports; and it is minted afresh when a sentence builds a new space. Inside it, the semanticist's distinctions survive only as a weak, unprivileged taxonomy, with belief no more separated from counterfactual than from any other class. The causal evidence favors a shared LM representational format across builders, while leaving the semantics of English, and belief opacity itself, to targeted linguistic tests.

# 7 Limitations and future work

Replication breadth differs by result, and the two most specific causal claims have the narrowest support. The representational (§4.3, 4.4), output-writing (§4.5), DAS causal-rank (§4.6), and geometric-taxonomy (§4.11) results hold on three families (Qwen, Pythia, Falcon3). The causal cross-type transfer (§4.8) holds on all three but is partial on Falcon3 (transfer index 0.711 [0.662, 0.741], the lowest of the three; Table 6). The report-vs-reasoning double dissociation (§4.7) holds on Qwen and Falcon3, the two families that can apply the sorting rule the test requires; Pythia-2.8b cannot (baseline consequent near chance), so it is uninterpretable there rather than a counterexample. The dissociation is backed by a measured inter-subspace angle (mean principal cosine 0.11 on Qwen, 0.17 on Falcon3), not the cross-effect asymmetry alone. As throughout, every causal number is the *margin* a learned subspace adds over the 0.58 leakage floor (Table 4), not a zero-baseline effect. All mechanistic results are on open models of 3B parameters or fewer, where activations are accessible; behaviorally the phenomenon also appears at frontier scale (gpt-4o, Qwen-2.5-72B, and Gemini-2.5 maintain both bindings, apply the rule, and hold eight coexisting spaces with no swap errors), so the small-model scope reflects intervention access, not a fragile effect.

We localize a distinct space index: low-rank, additive with entity identity, output-routed, causally compact, and shared across space types. Entity binding and belief tracking supply useful comparisons, but neither contains this variable. Entity binding keys entities or order slots; Prakash-style belief tracking keys character-object pairs to state values. Here the key is the discourse space. Whether this space index and entity binding are instances of a more abstract high-level causal model, in the sense of the causal-abstraction framework (Geiger et al., 2025), is left to future work. The present claim does not require that stronger identity.

Two threads remain open. The base vs image bindings are only *partially* separable: DAS-per-binding subspaces show the right directional asymmetry on Qwen and Falcon3 (each subspace moves its own value more than the other), but the leaky retrieval-shaped readout prevents a full double dissociation, which would need a model whose explicit-frame readout separates the frames more reliably. The reasoning task also needs to move beyond its single color-sorting rule, to rule out a template-artifact reading. Relatedly, the stimuli remain controlled rather than naturalistic: a mixed-frame battery shows the decodability results survive surface variation across all nine tested builder types (§4.1), and a painting-router causal control transfers across four surface frames (§4.10). Fully naturalistic text remains untested; a systematic behavioral map of inference-capability across models and rules, the cross-model face of the maintenance-vs-inference split (§4.2, 4.7, 4.9), is part of that same generalization beyond the single rule.

# 8 Data and code availability

A controlled stimulus suite for the mental-space constructions studied here is released as the Mental Spaces Corpus (Steele, 2026) (doi:10.57967/hf/9657 https://huggingface.co/datasets/osteele/mental-spaces). Build and validation code is at https://github.com/osteele/mental-spaces.